\documentclass{article}

\usepackage[preprint, nonatbib]{neurips_2023}

\usepackage{authblk}
\usepackage[utf8]{inputenc} 
\usepackage[T1]{fontenc}    
\usepackage{hyperref}       
\usepackage{url}            
\usepackage{booktabs}       
\usepackage{amsfonts}       
\usepackage{nicefrac}       
\usepackage{microtype}      
\usepackage{xcolor}         
\usepackage{amsmath}
\usepackage{amssymb}
\usepackage{mathtools}
\usepackage{amsthm}
\usepackage{multirow}
\usepackage{graphicx}

\usepackage{caption} 
\usepackage{makecell}

\usepackage{algorithmic}
\usepackage[linesnumbered,ruled,vlined]{algorithm2e}
\usepackage{cleveref} 
\usepackage{subcaption} 
\usepackage{float}

\usepackage{soul}

\DeclareMathOperator*{\argmin}{arg\,min}

\title{Discovering Galaxy Features via Dataset Distillation}

\author[1]{Haowen Guan}
\author[1]{Xuan Zhao}
\author[1]{Zishi Wang}
\author[1]{Zhiyang Li}
\author[1,2]{Julia Kempe}
\affil[1]{Center for Data Science, New York University}
\affil[2]{Courant Institute of Mathematical Sciences, New York University}

\begin{document}

\maketitle

\begin{abstract}
In many applications, Neural Nets (NNs) have classification performance on par or even exceeding human capacity. Moreover, it is likely that NNs leverage underlying features that might differ from those humans perceive to classify. Can we ``reverse-engineer'' pertinent features to enhance our scientific understanding? Here, we apply this idea to the notoriously difficult task of galaxy classification: NNs have reached high performance for this task, but what does a neural net (NN) ``see'' when it classifies galaxies? Are there morphological features that the human eye might overlook that could help with the task and provide new insights? Can we visualize tracers of early evolution, or additionally incorporated spectral data? We present a novel way to summarize and visualize galaxy morphology through the lens of neural networks, leveraging Dataset Distillation, a recent deep-learning methodology with the primary objective to distill knowledge from a large dataset and condense it into a compact synthetic dataset, such that a model trained on this synthetic dataset achieves performance comparable to a model trained on the full dataset. We curate a class-balanced, medium-size high-confidence version of the Galaxy Zoo 2 dataset, and proceed with  dataset distillation from our accurate NN-classifier to create synthesized prototypical images of galaxy morphological features, demonstrating its effectiveness. Of independent interest, we introduce a self-adaptive version of the state-of-the-art Matching Trajectory algorithm to automate the distillation process, and show enhanced performance on computer vision benchmarks.
\end{abstract}

\section{Introduction and Background}

The study of galaxy morphology is fundamental in observational cosmology. Morphological features are essential for determining a galaxy's dynamical state and interpreting its evolutionary history. Since Hubble’s first classification in 1926, significant efforts have been dedicated to designing morphological classification schemes and data collection methods. For instance, Galaxy Zoo \cite{gzoo, gzoo1}, through its crowd-sourcing approach for large-scale analysis, classifies galaxies from the Sloan Digital Sky Survey (SDSS) \cite{SDSS} into three basic types: elliptical (early-type), spiral (late-type), and mergers. Its successor, Galaxy Zoo 2 (GZ2, \cite{Willett_2013}), further expands this classification scheme to include more detailed morphological features, such as bars, bulges, and the shape of edge-on disks. Deep learning techniques, specifically those based on deep convolutional neural networks (CNNs, \cite{lecun-nature-15}), have emerged as automated approaches for galaxy morphology classification \cite{Dieleman_2015, Huertas_Company_2015}, yielding impressive results surpassing previous methods in predicting classifications made by humans. 
CNN-based galaxy morphology classification has now been applied across multiple different surveys, including SDSS \cite{Dom_nguez_S_nchez_2018, Walmsley_2019, Ghosh_2020}, CANDELS \cite{Huertas_Company_2018, Hausen_2020, Ghosh_2020}, and Dark Energy Survey \cite{cheng2020optimizing, cheng2021galaxy} with predicted features included in official surveys such as the new catalogue in SDSS-IV DR17 \cite{Dom_nguez_S_nchez_2021}. 

Visual morphologies are notoriously hard to classify, given the variability of the data (e.g.~sensitivity to red-shift). 
Prior approaches have aimed to find a set of parameters that correlate with the visual morphology of a galaxy. 
These traditionally include concentrations, asymmetries, clumpiness (or smoothness), Gini coefficient, moments of light etc. \cite{scarlata2007cosmos, Huertas_Company_2010, Peth_2016}. Unfortunately, the values of these
parameters strongly depend on the data quality and red-shift, as they overlook an enormous amount of information contained in the pixels themselves. Thus, these approaches only provide rough morphological classifications into 2 or 3 classes. 

\begin{figure}[t!]
    \vspace{-5mm}
    \centering
    \noindent\hspace*{-0.5cm}\includegraphics[width=0.8\textwidth]{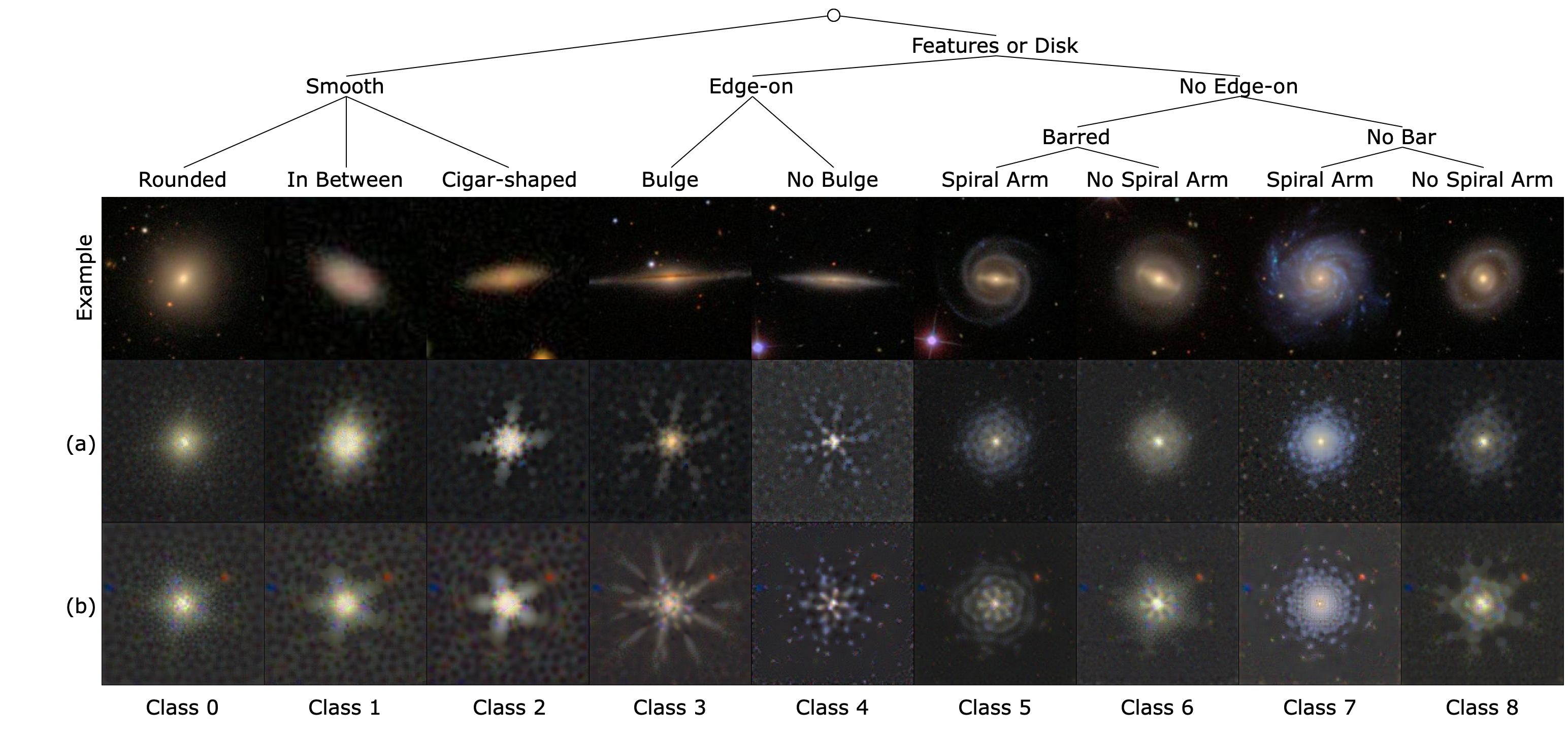}
    \vspace{-1mm}
    \caption{Classification tree based on Galaxy Zoo 2. The bottom two rows are STM-distilled images $128 \times 128$ (1 image per class): (a) starting from random real images, without rotational augmentation of the training set; (b) distilled from a rotationally augmented training set with synthetic data initialized from noise.}
    \label{fig:gz2_1ipc}
    \vspace{-4mm}
\end{figure}

We thus ask: Can the success of deep-learning based classification be leveraged to provide summary representations of morphological information {\em directly in the shape of galaxy images}? Such a transformation from successful CNN-based classifiers to synthesized summary images could then be extended to classifiers that process additional information (e.g. spectral data) for galaxy classification and generate images that are prototypical of morphology types even when additional non-visual measurement data is incorporated.
%
%
In this work we propose to leverage {\em Dataset Distillation (DD)} as a tool to achieve alternative summarization of galaxy morphologies in image form. DD, originally proposed by \cite{wang2020dataset} in computer vision, can be viewed as a form of dataset curation as a bi-level optimization task involving a neural net classifier. It aims to distill knowledge from a large dataset into a smaller one to reduce the burden of large-scale analysis on images. The dataset distillation optimization performs gradient descent on a synthesized dataset (outer loop) with respect to the loss (on real data) of a network trained on the distilled data (inner loop). Many directions have emerged from the initial bi-level optimization \cite{wang2020dataset}, including tractable approximations of the inner loop \cite{nguyen2020dataset, nguyen2021dataset, loo2022efficient, zhou2022dataset} and new objectives for optimization such as gradient matching \cite{pmlr-v139-zhao21a, zhao2021dataset}, trajectory matching \cite{cazenavette2022dataset, cui2022scaling}, distribution matching \cite{wang2022cafe, zhao2022dataset}
(see \cite{sachdeva2023data} for a recent survey).


Here, we focus on the Matching Training Trajectories (MTT) algorithm \cite{cazenavette2022dataset}, reaching recent state-of-the-art for various distillation benchmarks. We propose a new modification to MTT, called Self-Adaptive Trajectory Matching (STM) which allows for enhanced performance and ease on computer vision benchmarks. To apply it to galaxy distillation, we first curate an illustrative customized version of the GZ2 dataset as shown in Figure \ref{fig:gz2_1ipc}, and train a highly accurate CNN-based classifier on it. By condensing a considerable number of images into one or a few synthetic images for each category of galaxy, we can significantly reduce analysis time while revealing the essential morphological features for these categories. Dataset distillation emphasizes the features in the data that are essential to the classification: for instance, for our galaxy dataset, we will see that it enhances the blue features that are tracers of recent star formation in the galaxy. 

\section{Methodology and Data}

In the Dataset Distillation (DD) framework, the goal is to synthesize a compact dataset $\mathcal{D}_{syn}$ that can replicate the performance of a larger, real dataset $\mathcal{D} \in (\mathcal{X}, \mathcal{Y})$ when used with the same learning algorithm $\mathit{f}$. The optimal parameters for $\mathit{f}$ estimated on $\mathcal{D}$ and $\mathcal{D}_{syn}$ are represented as $\theta^{\mathcal{D}}$ and $\theta^{\mathcal{D}_{syn}}$, respectively. The objective of DD is to optimize: 

\begin{equation}\label{eq:DD_objective}
\underset{\mathcal{D}_{syn}}{\argmin} \left( |\mathit{f}_{\theta^{\mathcal{D}_{syn}}}(x) - \mathit{f}_{\theta^{\mathcal{D}}}(x) |  \;\;\; \forall x \in \mathcal{D} \right)
\end{equation}

This is an instance of a bilevel optimization problem where the output of one optimization ($\mathit{f}$ trained on ${\mathcal{D}_{syn}}$) is fed into another optimization problem (the generalization error on $\mathcal{D}$), which is computationally hard. 

\paragraph{Matching Training Trajectories (MTT):} The MTT method proposed by \cite{cazenavette2022dataset} aims to approximate the optimization in Equation (\ref{eq:DD_objective}) via gradient descent by minimizing the difference between $\theta^{\mathcal{D}}_{t}$ and $\theta^{\mathcal{D}_{syn}}_{t}$, for each iteration $t$ of weight updates during training. The objective is to identify a compact dataset $\mathcal{D}_{syn}$ such that when training on it, the model parameters $\theta^{\mathcal{D}_{syn}}$ closely resemble the teacher model parameters $\theta^{\mathcal{D}}$ when trained on the real dataset $\mathcal{D}$ throughout time. If we assume that the learning algorithm $\mathit{f}$ takes $T$ iterations to converge, the MTT objective can be formulated as:

\begin{equation}\label{eq:MTT_loss}
    \underset{\mathcal{D}_{syn}}{\argmin} \left(\|\theta^{\mathcal{D}_{syn}}_{t+N} - \theta^{\mathcal{D}}_{t+M}\|^2_2 / \|\theta^{\mathcal{D}}_{t} - \theta^{\mathcal{D}}_{t+M}\|_2^2 \;\;\; \forall t \in [0, \ldots, T) \right)
\end{equation}

In the equation above, $\theta^{\mathcal{D}}_{t}$ represents the model parameters for $\mathit{f}$ after $t$ iterations of training on dataset $\mathcal{D}$. The term $\|\theta^{\mathcal{D}}_{t} - \theta^{\mathcal{D}}_{t+M}\|_2^2$ serves as a normalization factor. While, in theory, the values of $N$ and $M$ should be equal (indicating a comparison after an equal number of gradient descent updates), in practice, especially when mini-batches are utilized for training and given $\mathcal{D} \gg \mathcal{D}_{syn}$, the ratio for $N:M$ becomes a hyperparameter.

MTT, while powerful, exhibits some shortcomings that stand in the way of scalability and practicality. i) MTT has a relatively large number of hyper-parameters that interact in a complex way: there are three parameters to control the trajectory matching mode and three different learning rates to configure, necessitating a significant amount of grid searching in hyper-parameter-space to achieve the optimal result. For instance, TESLA \cite{cui2022scaling}, a memory-optimized variant of MTT, outperforms MTT by varying hyper-parameter settings. ii) MTT lacks a clear stopping criterion for the distillation; the training process runs through a predetermined iterations. However, different datasets require different optimal iterations, and fixing the maximum iteration can lead to either excessive computation or sub-optimal training.

\setlength{\algomargin}{1em}
\begin{algorithm}[h]
    \caption{Self-Adaptive Trajectory Matching (STM)}
    \label{alg:STM}
    \KwIn{Teacher parameter trajectory set ${\Theta^{\mathcal{D}}}$, student network matching steps $N$, initial step size $\alpha$, threshold value for hypothesis test $\lambda$, maximum iterations per stage $\text{Max}_{iter}$}
    \textbf{Initialize:} $\mathcal{D}_{syn}$, $iter = 0$, $t = 0$, $T = 1$

    \While{$iter < \text{Max}_{iter}$}
    {
        Increment $iter$ and $t$ by 1, if $t$ reaches $T$, reset $t$ back to 0
        
        Sample $\theta^{\mathcal{D}}_{t}$ and $\theta^{\mathcal{D}}_{t + 1} \in \Theta^{\mathcal{D}}$ , set $\theta^{\mathcal{D}_{syn}}_t = \theta^{\mathcal{D}}_{t}$
        
        \For {$i = 1, ..., N$}
        {
            Update $\theta^{\mathcal{D}_{syn}}_{t + i} = \theta^{\mathcal{D}_{syn}}_{t + i - 1} - \alpha \nabla \ell (\theta^{\mathcal{D}_{syn}}_{t + i - 1}; \mathcal{D}_{syn})$
        }
        
        Update $\mathcal{D}_{syn}$ and $\alpha$ \footnotemark via gradient descent minimizing $\frac{\|\theta^{\mathcal{D}_{syn}}_{t+N} - \theta^{\mathcal{D}}_{t+1}\|_2^2 \;}{\; \|\theta^{\mathcal{D}}_{t} - \theta^{\mathcal{D}}_{t+1}\|_2^2}$   
        
        Repeat lines 4-6 replacing $t$ by $T$ to get $\theta^{\mathcal{D}_{syn}}_{T+N}$
        
        Collect validation loss $\frac{\|\theta^{\mathcal{D}_{syn}}_{T+N} - \theta^{\mathcal{D}}_{T+1}\|_2^2 \;}{\; \|\theta^{\mathcal{D}}_{T} - \theta^{\mathcal{D}}_{T+1}\|_2^2}$ into array $\ell_{val}$
        
        \tcc{Expand epoch pool if validation loss $\ell_{val}$ decreases fast enough}
        
        \If {$corr(\ell_{val}, time) < -\lambda \sqrt{1/(size(\ell_{val}) - 2)}$}
        {
            Expand epoch pool by increment $T$, Reset $iter, \ell_{val}$
        }
    }
    \KwOut{$\mathcal{D}_{syn}$}
\end{algorithm}
\footnotetext{As in MTT \cite{cazenavette2022dataset}, we make the step size $\alpha$ learnable. This adaptability enables the distillation algorithm to autonomously determine the optimal step size for aligning with the teacher trajectory.}

\paragraph{Self-Adaptive Trajectory Matching (STM):} To simplify and remedy some of these shortcomings, we propose \textbf{S}elf-adaptive \textbf{T}rajectory \textbf{M}atching (STM) that achieves two desiderata: it eliminates the need for $M$ and $T$ in \Cref{eq:MTT_loss}, and introduces an early stopping mechanism that can accurately halt the training process upon reaching the optimal result.

In vanilla trajectory matching algorithms, distillation involves selecting a maximum starting epoch $T$ and randomly sampling a starting point $t \in [0, \ldots, T)$ on the trajectory $\Theta^{\mathcal{D}} := \{\theta^{\mathcal{D}}_t\}^{T - 1}_{0}$ to proceed with parameter matching as in Equation \ref{eq:MTT_loss}. The $T$ can be interpreted as the size of trajectory $\Theta^{\mathcal{D}}$. MTT \cite{cazenavette2022dataset} demonstrates that the trajectory size $T$ and the distillation performance exhibit a parabolic relationship, and the optimal $T$ is positively correlated with the number of images per class (IPC) to distill for the synthetic dataset. An interpretation is that each teacher epoch $\theta^{\mathcal{D}}_t$ carries an amount of knowledge, and the synthetic dataset (student) has a certain ``capacity". If we can measure the capacity of a synthetic dataset during distillation, we can decide whether we should feed more teacher epochs to the student or decide to end the distillation.

To achieve this goal, we propose using a validation loss curve as an indicator of the capacity of the synthetic dataset. This validation loss is calculated by matching $\mathcal{D}_{syn}$ on a part of the trajectory outside the training trajectory $\Theta^{\mathcal{D}}$.  If the validation loss decreases in a statistically significant manner, we infer that the synthetic dataset possesses ``capacity for more knowledge.'' Consequently, we expand $\Theta^{\mathcal{D}}$ by adding $\theta^{\mathcal{D}}_T$ (or equivalently, by incrementing $T$). To determine statistically significant decreases in validation loss, we employ hypothesis testing \cite{wiki:Statistical_hypothesis_testing} on the correlation between validation losses (as time-series data) and time (distillation steps). Specifically, the null hypothesis is that the time-series data display an average zero correlation with time, and the deviation $\sigma$ is proportional to $\sqrt{1/(size(\ell_{val}) - 2)}$ (see Appendix \ref{tab:A1}). We establish a threshold of $\lambda \sigma$ to convincingly reject the null and expand training trajectory $\Theta^{\mathcal{D}}$ during distillation. If we cannot reject the null hypothesis after a certain maximum distillation steps, denoted as $\text{Max}_{iter}$, we stop. Furthermore, our findings suggest that fixing the teacher matching epoch to $M = 1$ is optimal, and cycling through $[0, \ldots, T)$ is better than randomly sampling from that interval. An ablation study on the parameter $\lambda$ is presented in Figure \ref{fig:sigma}, and shows that there is minimal variation when the value is sufficiently large. Our approach is summarized in Algorithm \ref{alg:STM}.

\textbf{Galaxy Zoo 2 (GZ2) and our curated GZ2:}
GZ2 \cite{Willett_2013} is renowned for its vast collection of 243,500 galaxies and the most reliable morphological classifications. In the GZ2 project, human volunteers are presented with galaxy images and are tasked with providing detailed descriptions of their morphologies by answering a series of  questions along a classification tree regarding its morphology, such as “Is the galaxy simply smooth and rounded, with no sign of a disk?” 
The GZ2 tree encompasses 11 classification tasks, with a total of 37 potential responses, leading to a vast number of possible classes with extreme class imbalance, with image counts ranging from 1,761 to 87,139. 
To partly mediate this, we simplify the classification tree to only 9 leaf nodes (classes), as illustrated in Figure \ref{fig:gz2_1ipc} by merging smaller, similar classes.
Each classification is labeled with a confidence determined by averaging across responses.
To assess the reliability of the dataset, we computed the average confidence level across the 9 classes to lie around 0.53, indicating sub-optimal data quality. To address this and restore class balance, we opted to select the top 600 most confidently classified galaxy images for each class for an average score of $0.79$, dividing them into 500 train and 100 test images per class. This ``higher-confidence" version of GZ allows for much higher training and test accuracies. For instance, our CNN classifier only achieves $56\%$ accuracy on the entire GZ dataset, while giving $89\%$ test accuracy on our curated dataset. Moreover the much smaller size of the curated data set enables active learning: astronomers can follow-up on these archetypes with additional spectroscopic and multi-frequency observations. Additionally, noting that data augmentation is often helpful in deep learning applications and given that galaxy imaging does not have a preferred orientation, we also construct an augmented curated GZ2 by rotating each galaxy image 36 degrees for 10 times, resulting in 45,000 train and 900 (non-augmented) test images.

\section{Experiments and Results}


\paragraph{Experimental Setup:} We deploy a simple 3-layer 128-width ConvNet \cite{gidaris2018dynamic}, which aligns with the previous DD benchmarks \cite{cui2022dcbench}. For data augmentation during the training of the teacher trajectory, we employ DSA \cite{pmlr-v139-zhao21a}. Like MTT \cite{cazenavette2022dataset}, we apply ZCA whitening on all benchmark datasets for image preprocessing. During the distillation process, $\mathcal{D}_{syn}$ can be initialized in two distinct ways: (1) Gaussian noise initialisation and (2) initialization with real images, randomly sampled from the original dataset. To assess performance, we train five separate networks from scratch on each distilled dataset and report test accuracy. For benchmarking, we create a non-synthetic baseline of size $|\mathcal{D}_{syn}|$ by randomly sampling the corresponding number of images from each class from the original dataset and using them to train networks in a manner consistent with the above approach. While most distillation hyper-parameters remain the same as MTT to ensure a fair comparison, certain parameters are adjusted due to modifications in the algorithm. Detailed hyperparameter settings can be found in Appendix \ref{tab:hyperparameter}.

\begin{table}[t]
    \centering
    \captionof{table}{Performance (test accuracy $\%$) of MTT and STM, trained on distilled data initialized from random real images.}
    \label{tab:cifar_results}
    \scriptsize
    \setlength{\tabcolsep}{2pt}
    \resizebox{0.7\linewidth}{!}{
    \begin{tabular}{cc|r|rr|c}
    \toprule
    & Img/Cls & \multicolumn{1}{c|}{Random} & \multicolumn{1}{c}{MTT} & \multicolumn{1}{c|}{Ours} & Full Dataset\\\hline
    \multirow{3}{*}{CIFAR-10} & 1 & 14.4 $\pm$ 2.0 & 46.3 $\pm$ 0.8\; & \bf{ 47.7 $\pm$ 0.1\;} & \multirow{3}{*}{84.8 $\pm$ 0.1} \\
    & 10 & 26.0 $\pm$ 1.2 & 65.3 $\pm$ 0.7\; & \bf{ 65.7 $\pm$ 0.2\;} & \\ 
    & 50 & 43.4 $\pm$ 1.0  & 71.6 $\pm$ 0.2\; & \bf{ 72.7 $\pm$ 0.2\;}& \\ \hline
    \multirow{2}{*}{CIFAR-100} & 1 & 4.2 $\pm$ 0.3 &  24.3 $\pm$ 0.3\; & \bf{24.4 $\pm$ 0.4\;} & \multirow{2}{*}{56.2 $\pm$ 0.3} \\ 
    & 10  & 14.6 $\pm$ 0.5 & 40.1 $\pm$ 0.4\; &   \bf{41.6 $\pm$ 0.3\;} \\ \bottomrule
    \end{tabular}}
\end{table}

\begin{table}[t]
    \vspace{-3mm}
    \centering
    \caption{Test accuracy of DD deployed on curated GZ2 dataset (with and without augmentation) with real and noisy initialization, as well as random baseline. 
    }
    \resizebox{1\linewidth}{!}{
    \begin{tabular}{c|c|ccc|ccc}
    \toprule
    \multirow{2}{*}{Img/Cls} & \multirow{2}{*}{Random} & \multicolumn{3}{c|}{curated GZ2} & \multicolumn{3}{c}{curated GZ2-Aug} \\ \cline{3-8}
    & & real & noise & Full Dataset & real & noise & Full Dataset \\ \hline
    1 & 27.4 $\pm$ 0.5 & 53.1 $\pm$ 1.7 & 54.6 $\pm$ 1.1 & \multirow{2}{*}{84.5 $\pm$ 0.6} & \bf{54.7 $\pm$ 0.7} & 54.6 $\pm$ 2.0 & \multirow{2}{*}{89.0 $\pm$ 0.5} \\ 
    10 & 46.0 $\pm$ 0.9 & 63.9 $\pm$ 1.0 & 62.4 $\pm$ 0.9 & & \bf{68.4 $\pm$ 1.2} & 67.4 $\pm$ 1.3 \\ \bottomrule
    \end{tabular}}
    \label{tab:GZ2_distillation}
    \vspace{-4mm}
\end{table}

\paragraph{Benchmarking STM:} We benchmarked our STM on CIFAR-10 and CIFAR-100 \cite{krizhevsky2009learning}, two key computer vision datasets, using 1/10/50 images per class (IPC) in Table \ref{tab:cifar_results}. We compare it with the original MTT \cite{cazenavette2022dataset} and a baseline model trained on the same number of randomly selected images from each class (Random). STM shows slight performance gains over MTT as it aims to streamline the data distillation process, enhance the algorithm's robustness, and enable it to consistently achieve optimal results. Figure \ref{fig:robustness} indeed shows faster convergence and much smaller variance between trials. We see that STM consistently yields improvement, both in performance and in convergence and stability, and we advocate to adapt MTT methods that are currently used to include the modifications brought by STM.

\noindent
\begin{minipage}[t]{0.49\linewidth}
    \begin{figure}[H]
        \centering
    \includegraphics[width=\textwidth]{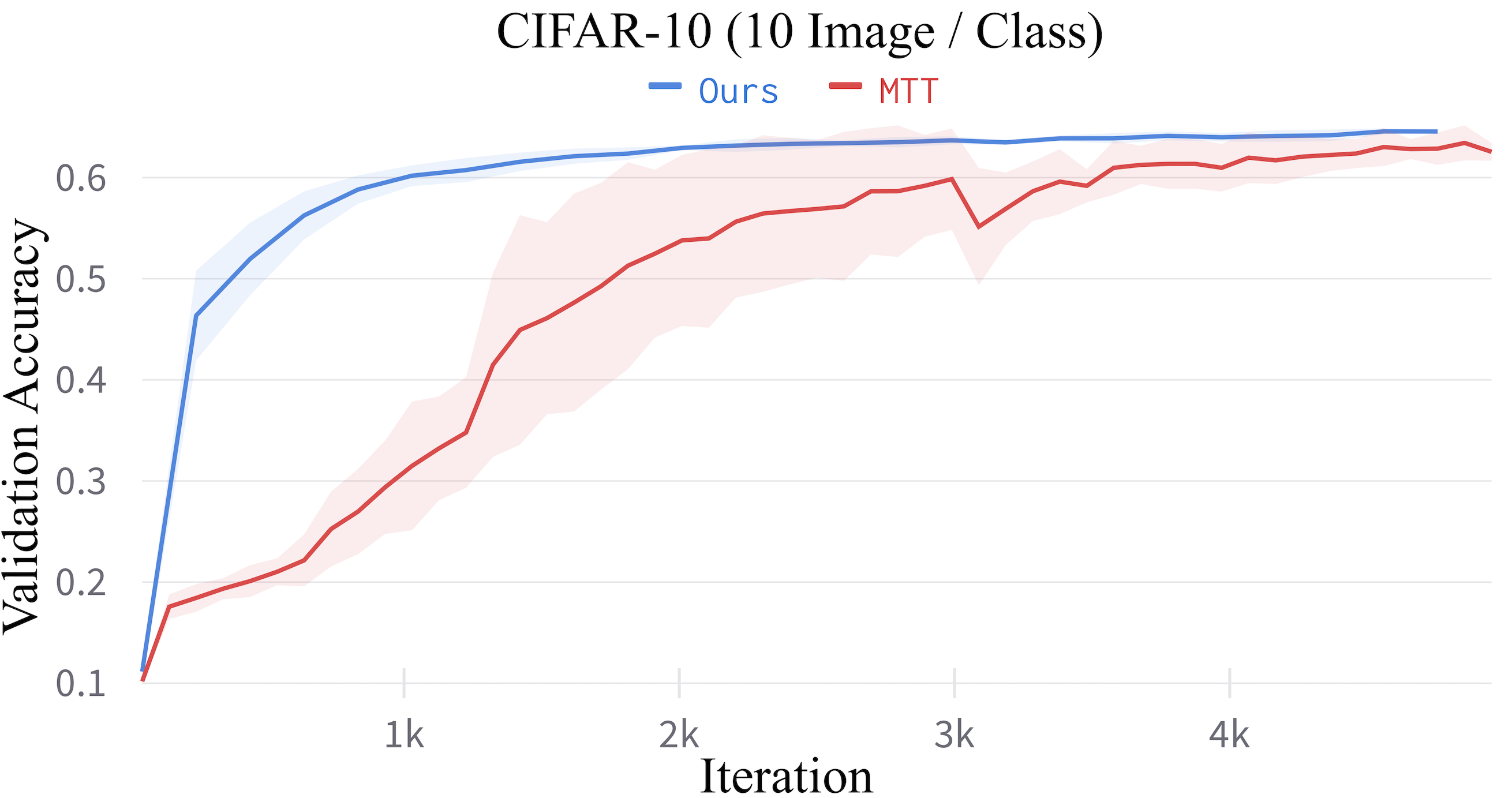}
    \caption{Comparison of MTT and STM distillation process (3 trials each, starting from random real images). STM shows a faster convergence speed and higher final accuracy.}
    \label{fig:robustness}
    \end{figure}
\end{minipage}
\hfill
\begin{minipage}[t]{0.49\linewidth}
    \begin{figure}[H]
        \centering
    \includegraphics[width=\textwidth]{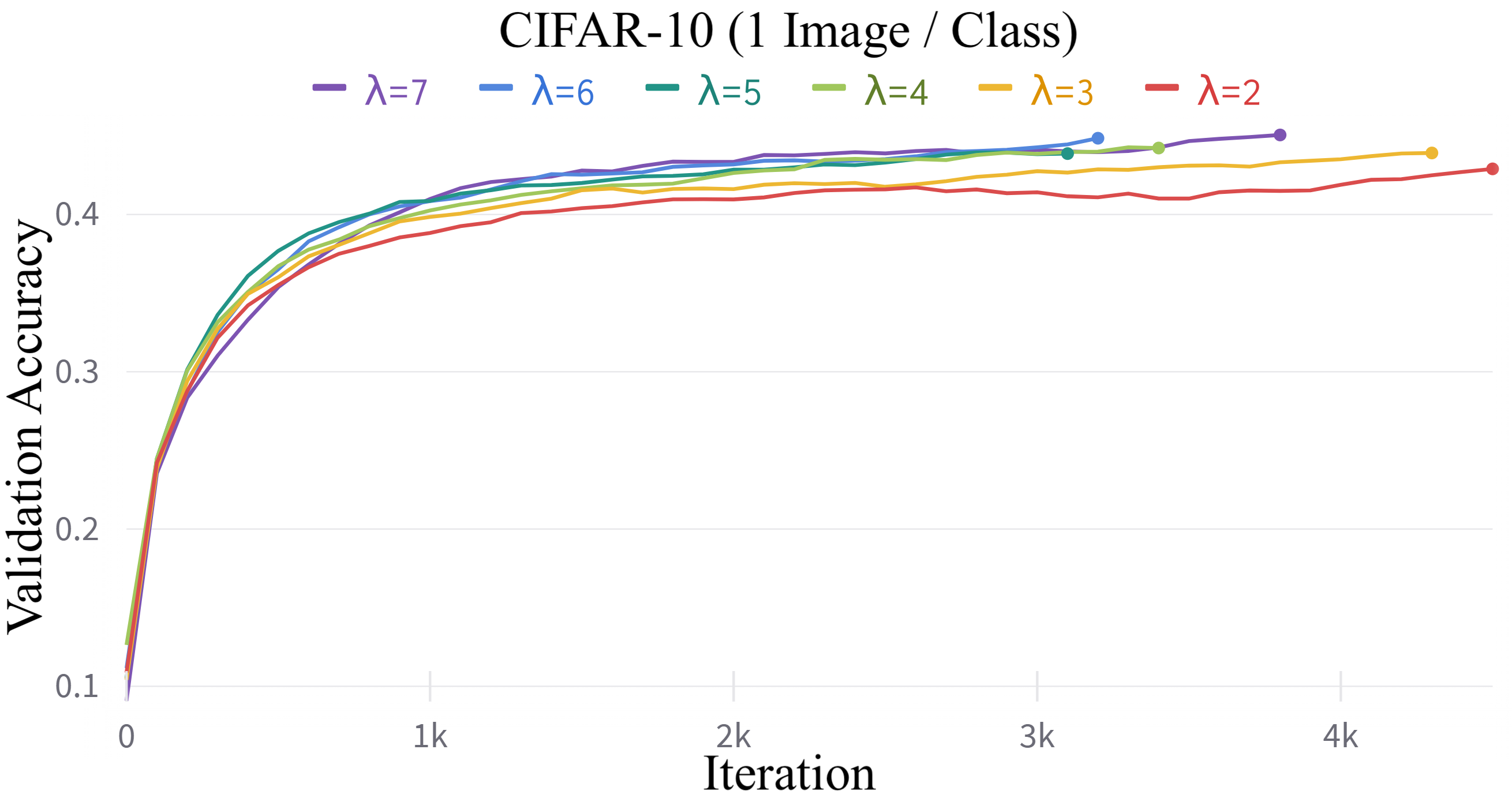}
    \caption{Ablation study for $\lambda$-sigma threshold of STM. A higher value implies a stricter requirement for statistical significance of decreasing loss trend. We fix our algorithm to use $5$-sigma.}
    \label{fig:sigma}
    \end{figure}
\end{minipage}

\begin{figure}[t]
    \centering
    \includegraphics[width=0.47\linewidth]{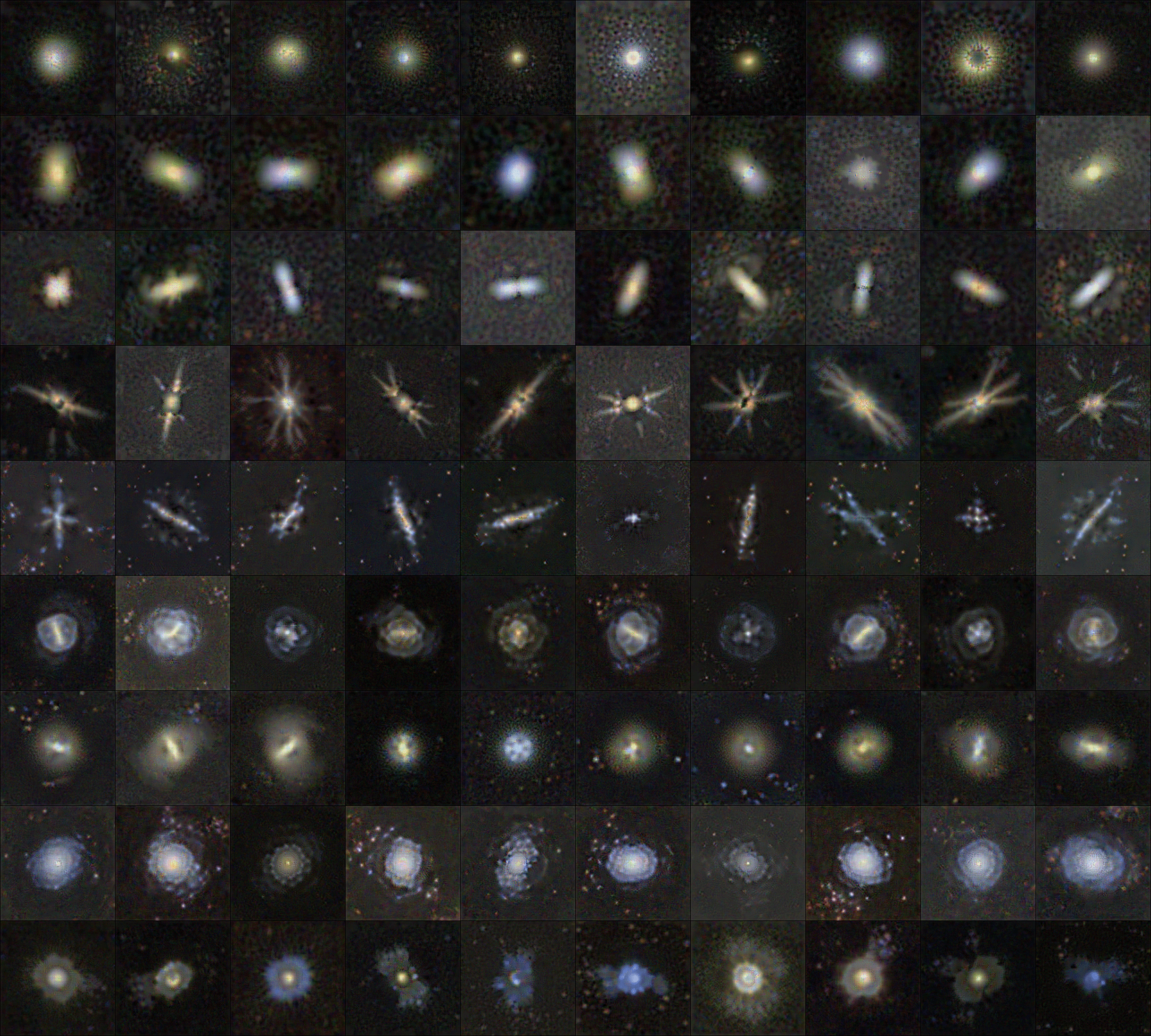}
    \hfill
    \includegraphics[width=0.47\linewidth]{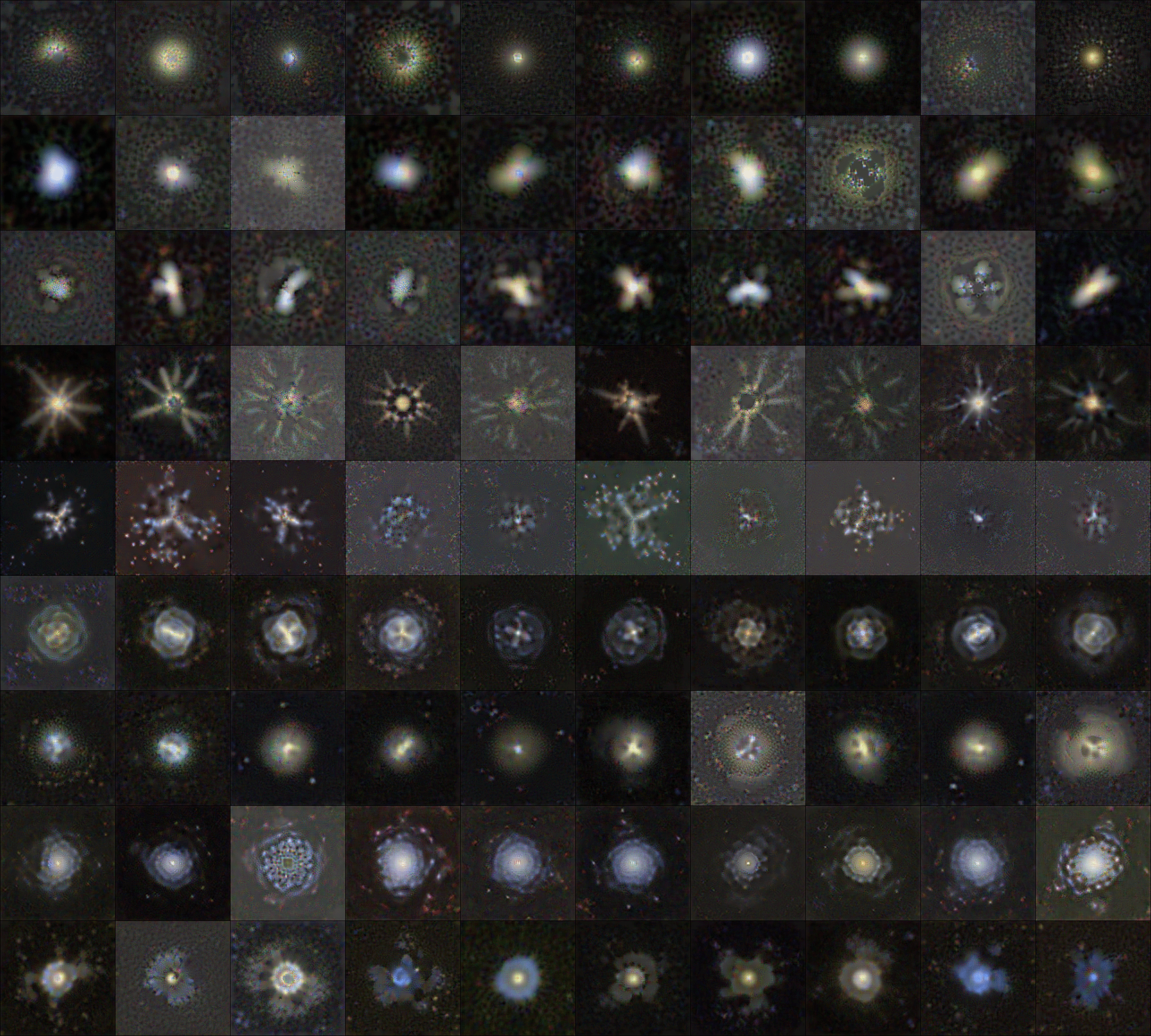}
    \caption{10 images per class distillation result for curated GZ2 with rotational augmentation; synthetic dataset initialized from real images (left) and noise initial (right). Each row belongs to one class ordered as in Figure \ref{fig:gz2_1ipc} (more images in the Appendix).}
    \label{fig:gz2_10ipc}
    \vspace{-3mm}
\end{figure}




\textbf{Distilling Galaxy Morphologies:} Table \ref{tab:GZ2_distillation} details the distillation results for both the basic and augmented curated GZ2 datasets, with 1 IPC examples shown in Figure \ref{fig:gz2_1ipc}. In our experiments, real synthetic data initialization (from random images) typically outperforms initialization from noise. While augmentation does not significantly improve 1 IPC accuracy, it boosts 10 IPC as well as accuracy when trained on the entire data by approximately $5\%$. Note that our CNN classifier yields a test-accuracy of $89\%$.

Distilled images capture the learned features of the model, providing insights into galaxy morphology that go beyond predefined characteristics, such as those used in the GZ2 survey questions. By optimizing classification performance, distilled images highlight key attributes for differentiation. The application of data distillation to galaxy imagery signifies the potential for leveraging machine-learned information to provide a fresh perspective and complement existing knowledge about galaxies.




For example, we can look at the 1 IPC distilled images in Figure \ref{fig:gz2_1ipc} (a). For Class 0-4 galaxies, key differentiators include core size and arm shape, taking into account the varying orientations of the galaxies. On the other hand, Classes 5\&7 exhibit a floral pattern due to asymmetrical arms; a smooth core for Class 7\&8 galaxies suggests the absence of a galactic bar. Distilled images provide a better understanding of a galaxy category’s overall characteristics. For instance, the blue tint of spiral galaxies indicates the presence of young stars. Compared to 1 IPC images, the 10 IPC versions retain more recognizable features, making them visually closer to real galaxies, as shown in Figure \ref{fig:gz2_10ipc}.



\section{Discussion}

We present a novel study that employs dataset distillation for the extraction of galaxy morphology features. We introduce a self-adaptive methodology, STM, which outperforms previous work 
and provide illustrative results to demonstrate that our approach is capable of distilling knowledge about galaxy morphologies, providing unique insights on key galaxy attributes that are not easily captured by human-based classification schemes. 
This opens the possibility of extracting even more informative images when adding spectroscopic or multi-frequency data as additional inputs to the classifier in the future. Our curated GZ2 dataset is of independent interest as an enabler for active learning of additional features on high-quality data. Our code is publicly available at \url{https://github.com/HaowenGuan/Galaxy-Dataset-Distillation}.

\section{Acknowledgements}

We gratefully acknowledge Adrian Price-Whelan, Marc Huertas-Company, and David Spergel for their pivotal physics insights and guidance regarding galaxy databases and potential insights to be had. Gratitude is also due to Jingtong Su for contributions to the dataset distillation algorithm. Helena Dominguez Sanchez’s assistance in preprocessing the galaxy dataset was invaluable. This work was supported by the National Science Foundation under NSF Award 1922658.

\medskip
{
\small
\bibliographystyle{plain}
\bibliography{main}
}


\newpage
\appendix
\section{Appendix}

\subsection{Standard Error of Correlation Coefficient Between Random Data and Time}
\label{tab:A1}

The correlation coefficient, denoted as $r$, quantifies the strength and direction of the linear relationship between two sequences of data with the same length $n$. The standard error $S_r$ of a correlation coefficient is given by (see e.g.~\cite{corr_book}):

\[S_r = \sqrt{\frac{1 - r^2}{n - 2}}\]

Consider a scenario where we sample $n$ data points from a normal distribution $\mathcal{N}(0, 1)$ and treat them as a time series. When computing the correlation coefficient of this data with respect to time, the true correlation coefficient is $r = 0$, since a randomly sampled dataset is expected to have no correlation with time. In this specific case, the standard error (or deviation $\sigma$) is:

\[\sigma = \sqrt{\frac{1}{n - 2}}\]
which is the functional form we use in \Cref{alg:STM}.

\subsection{Additional Hyperparameters}

\Cref{tab:hyperparameter} shows the hyperparameters we use for STM for various datasets. We set $\lambda=5$ in our hypothesis testing step but note that the algorithm is highly insensitive to the value of $\lambda$. We also have a parameter for maximum distillation steps, $\text{Max}_{iter}$; its value is fixed to $\text{Max}_{iter} = 1000$ in our experiment.

\begin{table}[h]
    \centering
    \vspace{-4mm}
    \captionof{table}{Hyper-parameters used for our best-performing distillation experiments. We adopt the terminology and definitions from MTT \cite{cazenavette2022dataset} (Pixel, Step Size). In the STM algorithm, $M$, the number of expert epochs is fixed to $1$.}
    \vspace{2mm}
    \label{tab:hyperparameter}
    \scriptsize
    \setlength{\tabcolsep}{2pt}
    \resizebox{0.9\linewidth}{!}{
    \begin{tabular}{cc|ccccc}
    \toprule
    dataset & Img/Cls & \multicolumn{1}{c}{\makecell{Synthetic Steps \\ ($N$)}} & \makecell{Learning Rate\\(Pixels)} & \makecell{Initial Step Size\\($\alpha$)} & \makecell{Learning Rate\\(Step Size)} & ZCA \\ \hline
    \multirow{3}{*}{CIFAR-10} & 1 & 50 & 1000 & 0.01 & 0.01 & Y \\
                              & 10 & 30 & 1000 & 0.01 & 0.01 & Y \\ 
                              & 50 & 30 & 1000 & 0.01 & 0.01 & Y \\ \hline
    \multirow{2}{*}{CIFAR-100} & 1 & 20 & 1000 & 0.01 & 0.01 & Y \\ 
                               & 10 & 20 & 1000 & 0.01 & 0.01 & Y \\ \hline
    \multirow{2}{*}{GZoo2} & 1 & 50 & 10000 & 0.0001 & 0.01 & N \\ 
                           & 10 & 20 & 10000 & 0.0001 & 0.01 & N \\ \bottomrule
    \end{tabular}}
\end{table}

\subsection{Additional Images}

Here we present additional visualizations of the distilled data from GalaxyZoo for various initializations (random or from noise) and distilled images per class.

\begin{figure}[h]
    \centering
    \includegraphics[width=0.48\linewidth]{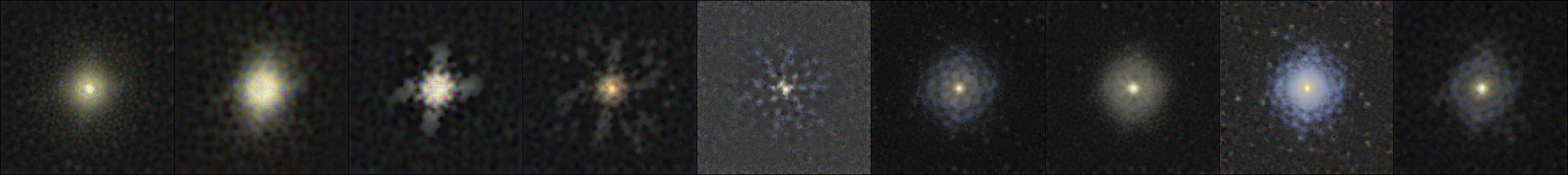}
    \includegraphics[width=0.48\linewidth]{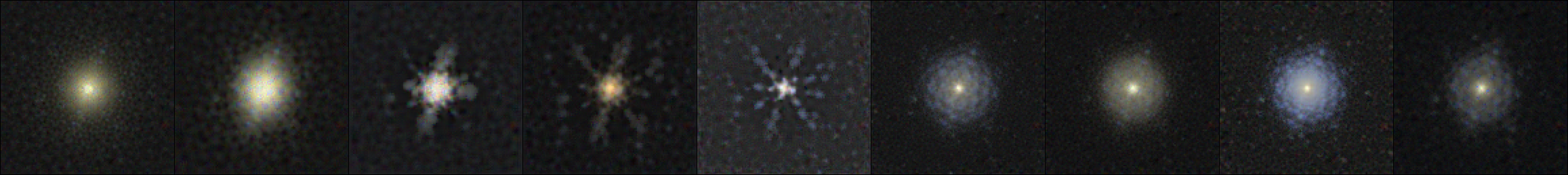}
    \caption{Distilled images 1 img/class without augmentation. Left: initialize from noise; Right: initialize from real}
    \label{fig:gz2_1ipc_noaug}
\end{figure}

\begin{figure}[h]
    \centering
    \includegraphics[width=0.48\linewidth]{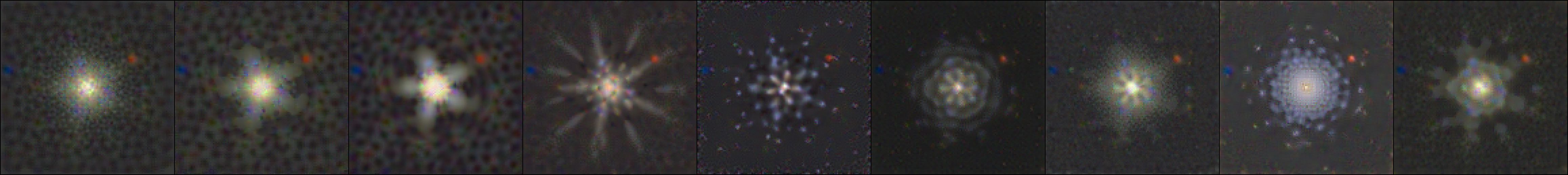}
    \includegraphics[width=0.48\linewidth]{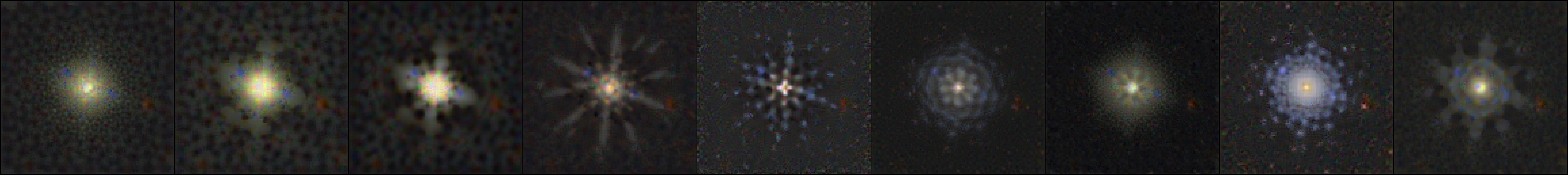}
    \caption{Distilled images 1 img/class with augmentation. Left: initialize from noise; Right: initialize from real}
    \label{fig:gz2_1ipc_aug}
\end{figure}

\begin{figure}[h]
    \centering
    \includegraphics[width=0.48\linewidth]{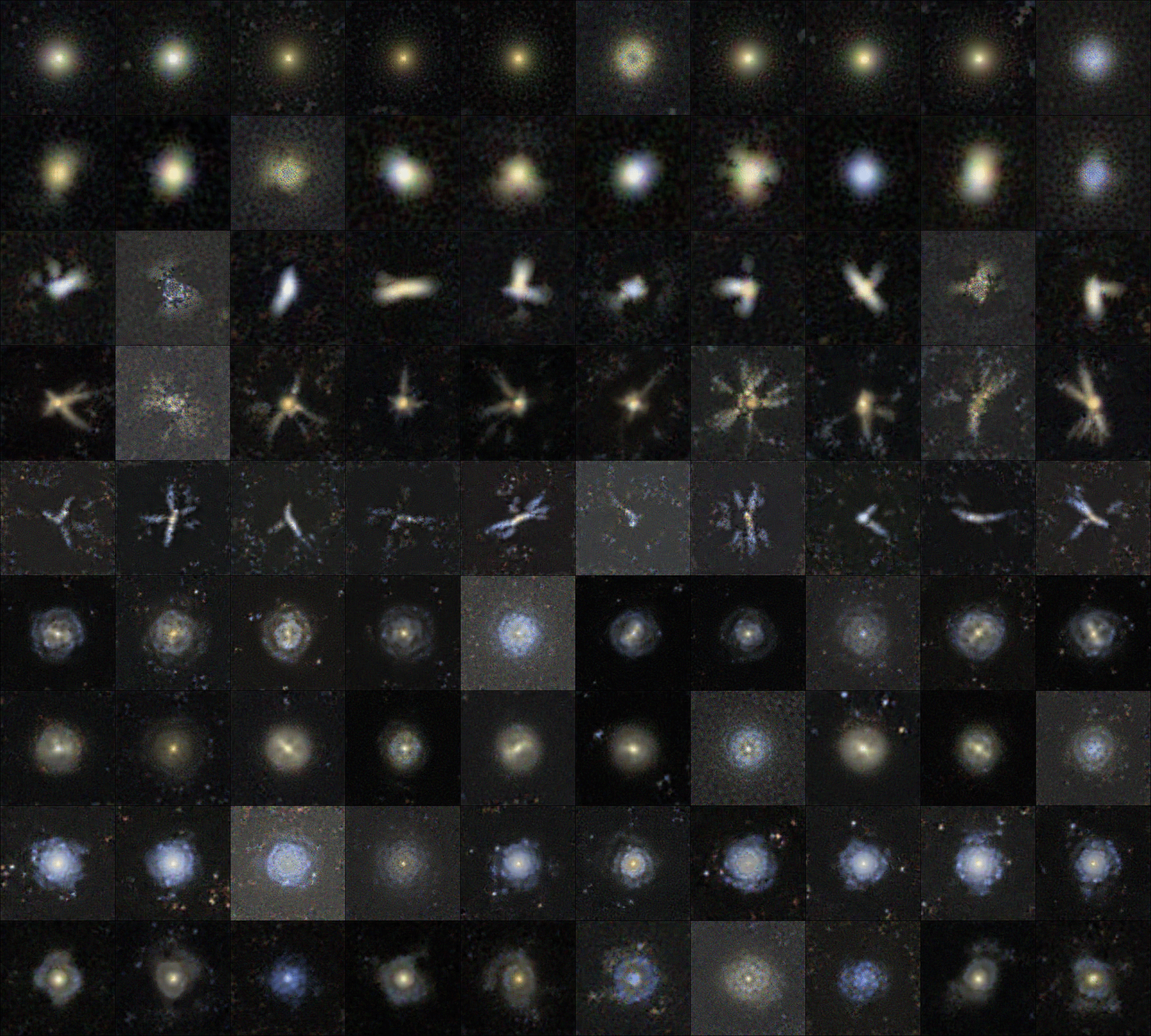}
    \includegraphics[width=0.48\linewidth]{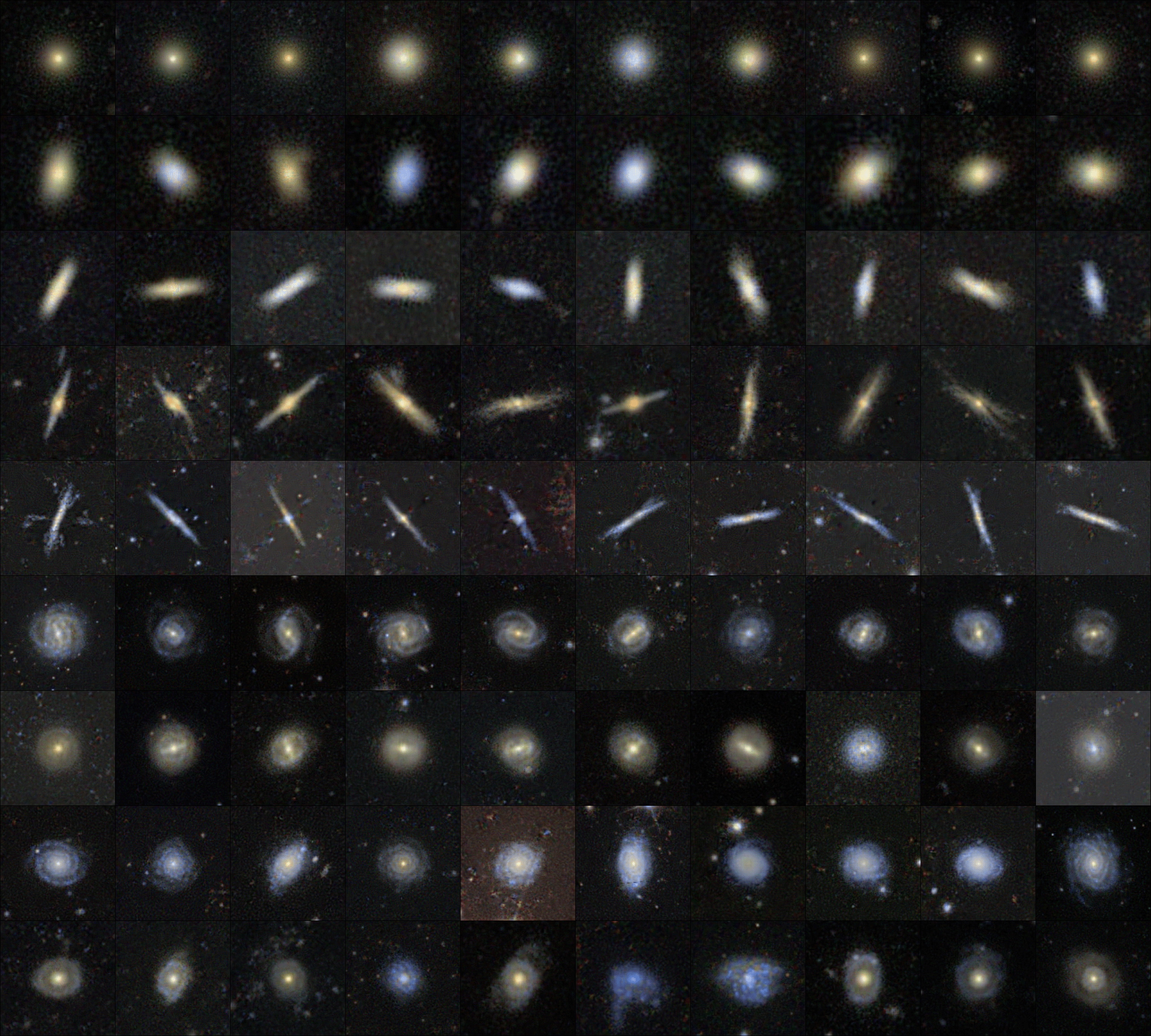}
    \caption{Distilled images 10 img/class without augmentation. Left: initialize from noise; Right: initialize from real}
    \label{fig:gz2_10ipc_noaug}
\end{figure}


\end{document}